\documentclass{article}
\usepackage{pgfplots}
\pgfplotsset{compat=1.18}
\usepackage{arxiv}
\usepackage{graphicx}   % For including images
\usepackage{hyperref}   % For PDF metadata and hyperlinks
\usepackage{ifpdf}      % To handle both LaTeX and PDFLaTeX
\usepackage{caption}    % Better figure captions
\usepackage{float}      % To control figure placement
\usepackage{amsmath}    % For cases environment and math alignment
\usepackage{booktabs}

\usepackage{pgfplots}
\usepackage{csvsimple}
\usepackage{subcaption}

\ifpdf
    \pdfoutput=1
\fi

\title{Moving Flock Detection in Pedestrian Trajectories Using Sequential Deep Learning Models}

\author{
Amartaivan Sanjjamts\\
Department of Information and Physical Sciences\\
Graduate School of Information Science and Technology\\
The University of Osaka, Osaka, Japan\\
\texttt{s.amaru@ist.osaka-u.ac.jp}\\
\And
Morita Hiroshi\\
Department of Information and Physical Sciences\\
Graduate School of Information Science and Technology\\
The University of Osaka, Osaka, Japan\\
\texttt{morita@ist.osaka-u.ac.jp}\\
\And
Enkhtogtokh Togootogtokh\\
Department of Research and Development\\
Voizzr Technologies\\
Bavaria, Germany\\
\texttt{enkhtogtokh.java@gmail.com}\\
}

\date{\today}

\begin{document}
\maketitle

\begin{abstract}
Understanding collective pedestrian movement is important for applications such as crowd analysis, space utilization, and the study of human mobility. This paper investigates sequential deep learning models, including Recurrent Neural Networks (RNNs), Long Short-Term Memory (LSTM) networks, and Transformers, for detecting moving flocks in pedestrian trajectory data. The proposed workflow has an offline preparation and analysis process: trajectory data are preprocessed and organized into scene sequences, a binary classifier is trained to distinguish flocking from non-flocking pedestrian pairs, and the trained classifier is subsequently applied to prepared scenes. Pairwise predictions are then aggregated to identify multi-agent flocks.

The experiments use real-world, group-annotated pedestrian trajectory data and evaluate the models across different sequence lengths and movement patterns. The study compares classification accuracy and computational runtime for the evaluated model configurations, and examines the use of pairwise predictions for identifying larger pedestrian groups. The workflow is designed for analysis of prepared trajectory data; it does not evaluate an end-to-end real-time system operating directly on a live pedestrian stream. 

\end{abstract}

\section{Introduction}

The analysis of pedestrian trajectories has become an essential aspect of understanding human mobility patterns in environments such as urban spaces, transportation systems, and public gatherings. In particular, identifying groups or "flocks" of pedestrians moving together over a period of time is an important task in the analysis of collective movement. A flock can be defined as a group of individuals whose movements are highly correlated over time, often indicating a shared goal or destination. Detecting such flocks is not only important for crowd management and safety but also for enhancing the effectiveness of autonomous systems, such as self-driving cars, and improving human-robot interaction.

Collective motion in trajectory data can be categorized into different formats, including \textbf{flocks}, \textbf{convoys}, and \textbf{swarms} \cite{wang2020big}. A \textbf{flock} is a set of agents moving together within a limited spatial region over a specific time interval. A \textbf{convoy} extends this definition by maintaining the same group structure over longer periods, making it more stable in dynamic environments. A \textbf{swarm} represents a more loosely connected group, where individuals exhibit similar movement patterns but do not necessarily maintain fixed spatial relationships. In this study, we focus on \textbf{moving flock detection}, where groups of pedestrians dynamically form and dissolve while moving together over short time intervals.

Traditional approaches to flock detection have largely relied on density-based methods, such as clustering algorithms and graph-based representations, to identify groupings based on spatial proximity or movement patterns. While effective in some scenarios, these methods often struggle in dynamic environments with changing densities, irregular movement, or occlusions.

In contrast, modern data-driven approaches, particularly deep learning models, offer flexible ways to represent temporal dependencies in trajectory data. Techniques such as Recurrent Neural Networks (RNNs), Long Short-Term Memory (LSTM) networks, and Transformers can learn sequential patterns that are relevant to identifying collective movement in multi-agent settings. When trained on suitable trajectory examples, these models can estimate whether pairs of pedestrians exhibit movement patterns associated with flocking.

This paper investigates sequential deep learning models for moving-flock detection in pedestrian trajectory data. It compares RNN, LSTM, and Transformer models for pairwise classification and applies a trained pairwise classifier to prepared multi-pedestrian scenes. The resulting pairwise predictions are aggregated to estimate flock membership and size. The study evaluates this workflow on prepared, recorded trajectory data rather than a live, continuously arriving stream.

\section{Literature Review}

Understanding collective pedestrian movement has been an active area of research, with various methodologies proposed to detect and analyze groups of individuals moving together. These approaches can broadly be categorized into traditional model-based methods and modern data-driven techniques, with recent advancements exploring hybrid strategies to improve detection accuracy.

\subsection{Traditional Approaches}

Early research in flock detection primarily relied on model-based and clustering techniques. One of the most widely used methods is Density-Based Spatial Clustering of Applications with Noise (DBSCAN), which identifies pedestrian groups based on spatial proximity and density \cite{ester1996density}. While effective in many cases, DBSCAN and other density-based techniques face challenges in dynamic environments where pedestrian group sizes and movement patterns change over time.

Graph-based methods have also been extensively used, where pedestrians are represented as nodes and edges denote spatial or temporal interactions \cite{helbing2000social}. These methods often employ social force models to simulate pedestrian dynamics, considering factors like inter-agent distance and collision avoidance. While useful for understanding local interactions, such approaches struggle with long-range dependencies and occlusions, limiting their applicability in real-world, large-scale crowd scenarios.

Another line of research involves rule-based and statistical models, such as Hidden Markov Models (HMMs) and Kalman filters, to estimate pedestrian movement patterns. These models typically assume linear motion patterns and rely on handcrafted features, making them less adaptable to complex and unstructured pedestrian flows \cite{li2008learning}. 

\subsection{Modern Data-Driven Approaches}

With the rise of deep learning, trajectory-based modeling has shifted towards data-driven methods that learn patterns directly from large-scale pedestrian datasets. Recurrent Neural Networks (RNNs) and Long Short-Term Memory (LSTM) networks have been widely used for pedestrian trajectory prediction and group detection due to their ability to capture sequential dependencies \cite{zhang2016deep}. These models have demonstrated improved accuracy in detecting pedestrian flocks by learning motion tendencies over time.

More recently, Transformer-based architectures have emerged as state-of-the-art models in sequence learning tasks. Unlike RNNs and LSTMs, Transformers rely on self-attention mechanisms, allowing them to model long-range dependencies more effectively without suffering from vanishing gradient issues \cite{vaswani2017attention}. This has led to improved performance in large-scale pedestrian behavior analysis, where long-term dependencies and complex interactions are prevalent.

Additionally, attention-based models, such as Social-LSTM and Social-GAN, have been proposed to model social interactions among pedestrians, dynamically adapting their predictions based on surrounding agents \cite{gupta2018social}. These models leverage learned representations to classify pedestrian groups, offering a more flexible alternative to traditional rule-based approaches.

\subsection{Hybrid Approaches}

Recent studies have explored the integration of traditional and deep learning-based methods to enhance flock detection capabilities. Hybrid models, such as DBSCAN combined with LSTMs, have been used to detect groups first through clustering and then refine predictions using sequential learning models \cite{jia2020trajectory}. Similarly, Graph Neural Networks (GNNs) have been applied to pedestrian movement analysis, capturing both spatial and temporal dependencies by integrating graph structures with deep learning architectures \cite{zhang2020graph}.

The combination of these approaches has demonstrated potential in dynamic and large-scale scenarios where traditional clustering methods alone may be insufficient. By leveraging spatial relationships and deep sequence modeling, hybrid models provide a framework for multi-agent flock analysis.

\section{Methodology}

In this study, we leverage sequential deep learning models to classify trajectory pairs as either forming a flock or not. The task is based on the premise that identifying patterns of movement within groups of pedestrians can offer valuable insights into pedestrian dynamics, such as crowd behavior, space utilization, and evacuation strategies. The problem is framed as a binary classification task, where each trajectory pair is classified as belonging to a flock (pair) or not (non-pair). 

The scope of this study is offline analysis of recorded trajectory data. Trajectory preprocessing, time-bin grouping, scene-sequence generation, and model training are performed as preparatory stages. The trained model is then applied to the prepared scene sequences for pairwise classification, after which the predictions are aggregated into multi-agent flocks. The study does not implement or evaluate continuous online processing of live trajectory streams, nor does it report end-to-end real-time latency.

In this context, a \textbf{flock} is formally defined as a subset of pedestrians whose trajectories exhibit sustained spatial proximity and coherent motion patterns over a contiguous temporal interval. The spatial constraint is characterized by bounded inter-pedestrian distances and consistent relative orientations, while the temporal constraint requires persistence of these spatial relationships for a minimum duration. Flock detection thus involves the extraction of \textbf{spatial features}---such as Euclidean inter-pedestrian distance, relative velocity magnitude, differences in motion and facing angles, and trajectory similarity metrics (e.g., Dynamic Time Warping)---and \textbf{temporal features}---such as absolute timestamp differences, duration of spatial cohesion, and stability of motion direction. These spatio-temporal descriptors collectively enable the model to discriminate between coordinated group movement and incidental co-location arising from independent pedestrian motions.

To achieve this, we employ a variety of sequential deep learning models. Specifically, we focus on architectures such as Recurrent Neural Networks (RNNs), Long Short-Term Memory (LSTM) networks, and Transformer-based models. Each of these models is trained to predict pairwise relationships between pedestrians based on their movement characteristics. The core idea is to capture temporal dependencies in the movement data, allowing the models to understand how pairs of pedestrians interact over time and how these interactions evolve.

\begin{figure}[H] 
    \centering
    \ifpdf
        \includegraphics[width=0.8\textwidth]{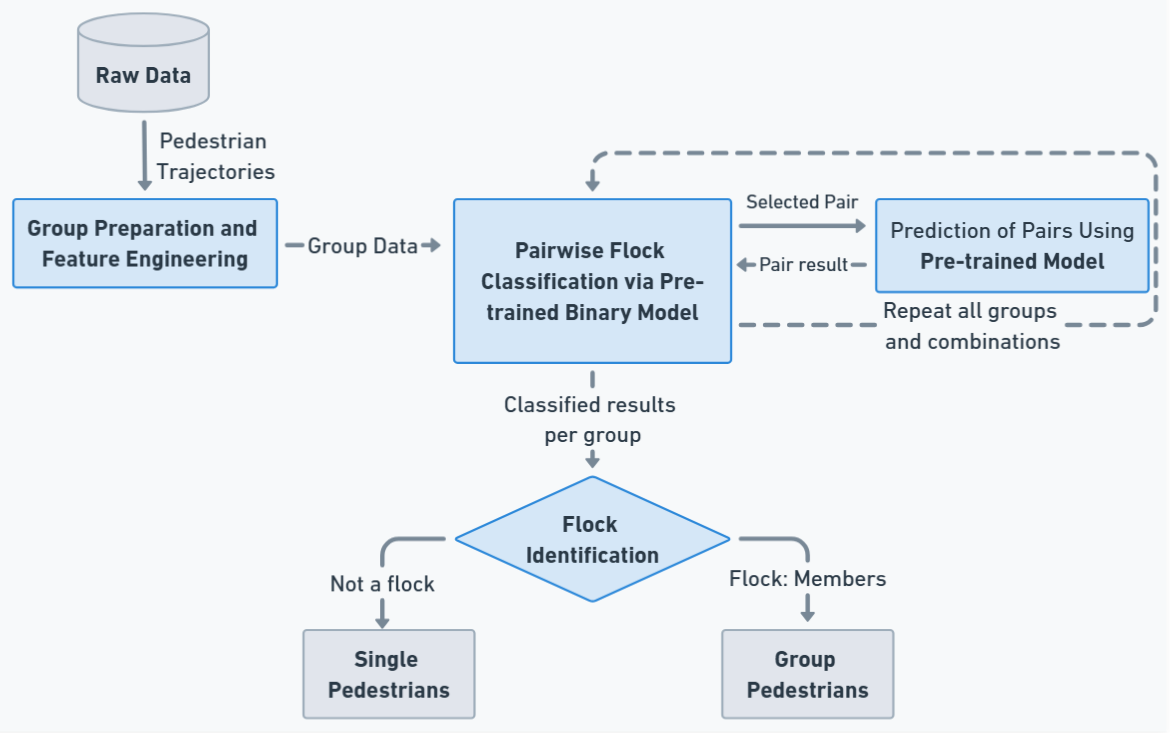}  % Use PDFLaTeX-compatible formats
    \fi
    \caption{Proposed Method Overview}
    \label{fig:proposed_method}
\end{figure}

As shown in Figure \ref{fig:proposed_method}, the proposed workflow consists of offline data preparation and model training, followed by analysis of prepared scenes. The main analysis steps are as follows:

\begin{itemize} \item \textbf{Binary Classification:} We train a deep learning model specifically designed to predict whether a given pair of pedestrians belongs to a flock or not. This model is trained using trajectory data, where each pedestrian's movement features such as position, velocity, motion angle, and inter-agent distance are considered to make the classification decision. The binary classification framework allows the model to focus on detecting pairwise interactions, essential for identifying potential group behaviors. \end{itemize}

\begin{itemize} \item \textbf{Pairwise Evaluation:} After training, the classifier is applied to the pedestrian pairs represented in each prepared scene. Pairwise classification estimates whether each pair exhibits the spatio-temporal characteristics of flocking. This converts multi-agent group analysis into a set of binary classification decisions and supports scenes containing different numbers of pedestrians. In the experiments reported here, this evaluation is performed on prepared trajectory sequences, not as streaming inference on live sensor data. \end{itemize}

\begin{itemize} \item \textbf{Flock Filtering:} After the pairwise evaluations are completed, a filtering mechanism is applied to refine the detected flocks. This process involves using predefined criteria such as a minimum flock size (e.g., a minimum of two pedestrians) and temporal consistency (e.g., ensuring that pairs that are detected as a flock maintain their relationship over time). Flock filtering ensures that the detected groups represent meaningful pedestrian formations rather than isolated or spurious pairings. The filtering mechanism also helps in removing false positives and reduces the likelihood of detecting non-flocking pairs, thereby improving the accuracy of the model's output. \end{itemize}

\begin{itemize} \item \textbf{Flock Aggregation:} After filtering, the final step involves aggregating the results of pairwise evaluations into larger flock formations. By clustering pairs that are consistently detected as belonging to the same flock, we can estimate the size and structure of the group. This aggregation step is key in translating the pairwise predictions into actionable insights, allowing us to identify the overall structure of pedestrian groups, such as convoys, clusters, or swarms, in the scene. The resulting groups are then analyzed for their behavior, density, and movement patterns. \end{itemize}

\subsection{Binary Classification Problem}

We formulate pairwise moving-flock identification as a binary classification problem. Given a pair of pedestrian trajectory sequences from a prepared scene, the goal is to determine whether the two pedestrians belong to the same flock over the represented observation interval.  

Let $\mathbf{T}_i = \{(x_t^i, y_t^i) \mid t = 1, \dots, T\}$ and $\mathbf{T}_j = \{(x_t^j, y_t^j) \mid t = 1, \dots, T\}$ represent the trajectories of pedestrians $i$ and $j$, where $(x_t, y_t)$ denotes the position of a pedestrian at timestamp $t$.  

We define a binary function $\mathcal{F}$ to classify whether the pair $(i, j)$ belongs to the same flock:  

\begin{equation}
    \mathcal{F}(\mathbf{T}_i, \mathbf{T}_j) =
    \begin{cases}
      1, & \text{if } (i, j) \text{ belong to the same flock}, \\
      0, & \text{otherwise}.
    \end{cases}
\end{equation}

To learn this function, we train a deep learning model $\mathcal{M}_{\theta}$ with parameters $\theta$ such that:  

\begin{equation}
    \hat{y}_{i,j} = \mathcal{M}_{\theta}(\mathbf{T}_i, \mathbf{T}_j),
\end{equation}

where $\hat{y}_{i,j} \in \{0,1\}$ represents the predicted label indicating whether the pair belongs to a flock.
\subsection{Group Data Preparation}\label{Group_Data_Preperation}

As an offline preprocessing stage, we organize recorded pedestrian trajectories by their starting points, under the assumption that pedestrians who start moving at similar times may belong to the same flock. This stage is completed before model-based scene analysis. The preparation process consists of two main steps: time-bin classification and sequence data selection.

\subsubsection{Time-Bin Classification}
First, the function read the preprocessed pedestrian trajectory data, which includes fields for timestamp, agent ID, position $(X,Y)$, velocity, motion angle, and face angle. To ensure uniform timestamp representation, the timestamps are converted into datetime format and adjusted to the appropriate time zone.

Next, we filter out agents with fewer trajectory points than the required sequence length ($L$). For each remaining agent, we identify the first recorded trajectory point (i.e., the minimum timestamp for each agent). These first trajectory points are then grouped into discrete time bins of size $T$, using:

\begin{equation}
\text{time\_bin} = \left\lfloor \frac{\text{TIMESTAMP} - \text{min}(\text{TIMESTAMP})}{T} \right\rfloor.
\end{equation}

Each bin represents a cohort of agents that started moving within the same time interval. The number of agents in each bin is recorded, and bins with zero agents are excluded. The resulting groups of agent IDs are stored in a file for further processing.

\subsubsection{Sequential Data Selection}
Once the agents are grouped into time bins, the trajectory data for each agent is extracted to form sequences of length $L$. For each agent, we ensure that a sufficient number of consecutive data points exist to meet the sequence length requirement. The data is then sorted by timestamp and agent ID.

The trajectory sequences for all agents within a given time bin are structured into blocks, where each block contains multiple agents' trajectories for a shared time interval. These structured blocks of sequential data are then exported as scene data for further analysis.

This preprocessing method organizes recorded trajectories into scene sequences based on agents' starting times. The resulting scene data are prepared before model-based pairwise evaluation and support subsequent analysis of collective movement.

\subsection{Multi-agent flock identification}\label{multi-agent}

To identify multi-agent flocks from pairwise classification results, we employ a union-find-based clustering approach. This method efficiently groups pedestrians into cohesive flocks based on their classified pairwise relationships. The algorithm processes an input list of pedestrian pairs that have been determined to be part of the same flock and iteratively merges them into larger clusters. By leveraging the union-find data structure, we ensure that the clustering process remains computationally efficient, even for large-scale pedestrian datasets. The final output is a set of detected flocks, where each flock represents a cohesive group of individuals moving together within the observed environment.

\subsubsection{Input and Output}

\begin{itemize}
    \item \textbf{Input:} A list of tuples, where each tuple contains:
    \begin{itemize}
        \item A pair of pedestrian identifiers $(p_1, p_2)$.
        \item A binary classification result: 1 if the pair belongs to the same flock, 0 otherwise.
    \end{itemize}
    \item \textbf{Output:} A list of detected flocks, each represented as:
    \begin{itemize}
        \item The number of pedestrians in the flock.
        \item A sorted list of pedestrian identifiers forming the flock.
    \end{itemize}
\end{itemize}

\subsubsection{Flock Detection Workflow}

The flock identification operates as follows:

\begin{enumerate}
    \item \textbf{Union-Find Initialization:} Each pedestrian is initially considered its own flock.
    \item \textbf{Pairwise Flock Merging:} For each detected flock pair:
    \begin{enumerate}
        \item Find the representative (root) of each pedestrian's flock using the \textit{find\_root} function.
        \item If the pedestrians belong to different flocks, merge them by updating their root.
    \end{enumerate}
    \item \textbf{Flock Organization:} After processing all pairs, pedestrians belonging to the same root are grouped into final flock structures.
    \item \textbf{Output Generation:} The function returns a list where each detected flock is represented by its size and the sorted list of pedestrian identifiers.
\end{enumerate}

\subsubsection{Union-Find with Path Compression}

To efficiently manage flock merging, we implement a union-find data structure with path compression. Given a pedestrian $p$, the function \texttt{find\_root(p)} recursively finds the root representative:

\begin{equation}
    \text{flocks}[p] =
    \begin{cases}
      p, & \text{if } p \text{ is its own root}, \\
      \text{find\_root}(\text{flocks}[p]), & \text{otherwise}.
    \end{cases}
\end{equation}

Path compression ensures that subsequent searches for the same pedestrian are efficient, reducing the overall complexity of the algorithm to nearly $\mathcal{O}(N)$ in practice.

\section{Experiments and Results}

We compare RNN, LSTM, and Transformer models in terms of accuracy and performance.

\subsection{Data and Model Preperation}

\subsubsection{Dataset}

We use a real-world pedestrian group-identified trajectory dataset \cite{zanlungo2015pedestrian} from an indoor environment recorded at the Asian Trade Center in Osaka, Japan.

Each row in the CSV file corresponds to a single tracked individual at a specific time instant and includes the following fields: \textit{time [ms] (Unix time + milliseconds/1000)}, \textit{person ID}, \textit{position x [mm]}, \textit{position y [mm]}, \textit{velocity [mm/s]}, \textit{angle of motion [rad]}, and \textit{facing angle [rad]}.

The group files contain annotations for the groups present on a given day. Only pedestrians in groups are listed, while those walking alone are excluded. Each row represents a single tracked pedestrian within a group and includes the following fields (space-separated values): $PEDESTRIAN-ID$, \textit{$GROUP-SIZE$}, \textit{$PARTNER-ID-1$}... (list of IDs of all other pedestrians in the group), \textit{$NUMBER-OF-INTERACTING-PARTNERS$}, and \textit{$INTERACTING-PARTNER-ID-1$}... (list of all socially interacting partners).

For binary-classifier training, we select data where \textit{$GROUP-SIZE$} = 2, focusing on pedestrian pairs from a single day. After training, the model is evaluated on data from different days that were not used for training. Thus, model training and the preparation of evaluation scenes are completed before the pairwise evaluation of those scenes.

\subsubsection*{Addressing Class Imbalance}

One challenge in training the binary classification model is class imbalance, as natural pedestrian movement data may contain significantly more non-flocking samples than flocking ones. To mitigate this issue, we apply a combination of:

Weighted Loss Functions: Assigning higher penalties to misclassified minority class samples.
Oversampling \& Undersampling: Augmenting minority class sequences and reducing redundant majority class samples.
Synthetic Data Generation: Creating additional flocking sequences by interpolating existing trajectories.
By employing these techniques, we ensure that the model does not develop a bias toward the majority class and maintains high predictive accuracy across both categories.
We split the dataset into training and testing sets with an 80/20 ratio. Depending on the sequence length (the number of records in a single sample), the number of samples vary.

\begin{table}[H]
   
    \label{tab:data_summary}
    \centering
    
    \begin{tabular}{@{}cccc@{}} 
     \toprule
     \textbf{Sequence Length} & \textbf{\# of total samples} & 
     \textbf{\# of training samples} & \textbf{\# of excluded agents} \\
     \midrule
    30 & 2794 & 2235  & 102 \\
    60 & 2758 &  2206 &  138\\
    100 & 2684 & 2147 &  212\\
    150 & 2612 & 2090 &  284\\
    200 & 2542 & 2034 & 354\\
    300 & 2404 & 1923 & 492\\
    500 & 2022 & 1618 &  874\\
     \bottomrule
    \end{tabular}  
    \caption{Sample Sequence Length vs. Number of Samples on 2013/05/05}
    \label{tab:seq_vs_sample}
\end{table}

As shown in Table \ref{tab:seq_vs_sample}, we extract sequence lengths of 30, 60, 100, 150, 200, 300, and 500. Based on the prepared pair pedestrian IDs, we select trajectory data where the current pedestrian meets the required sequence length, with a label of '1' indicating a flock. We then randomly select two single pedestrian datasets and add them to the dataset, labeling them '0' to indicate they are not part of a flock. As a result, the \textit{\# of total samples} include both pair and non-pair data, maintaining a 50-50 ratio.

\subsubsection{Deep Learning Model Preparation}

\subsubsection*{Feature Engineering}

In accordance with the definition of a flock, we extract features that capture spatial and behavioral interactions between pedestrians. These include the inter-distance between pedestrians, and absolute differences in motion properties such as velocity, motion angle, and facing angle. Additionally, we incorporate trajectory-based similarity metrics like Dynamic Time Warping (DTW) to enhance the model's ability to distinguish between temporary co-movement and sustained flocking behavior.

To ensure robust performance, we preprocess the extracted features through normalization techniques, such as min-max scaling and standardization, preventing bias due to differences in measurement units. Furthermore, we explore additional features related to pedestrian interaction, such as relative velocity angle, trajectory curvature, and acceleration differences.

For our binary classification sequential model, each sample consists of multiple sequential data points, where each instance corresponds to a label of '1' (indicating that the pair forms a flock) or '0' (indicating that they do not). The final set of features, selected based on their predictive importance, is listed in Table \ref{tab:feature_table}

\textbf{Feature Scaling:}  
To ensure numerical stability and improve model convergence, each feature is normalized using a specific scaler based on its distribution and sensitivity to outliers:
\begin{itemize}
\item  \textbf{RobustScaler} for \texttt{interDistance}: Reduces the influence of extreme values by scaling based on the median and interquartile range.  
\item \textbf{MinMaxScaler} for \texttt{timeDifference}: Normalizes values to a fixed range [0, 1], preserving temporal proportionality.  
\item  \textbf{StandardScaler} for \texttt{velocityDifference}, \texttt{motionAngleDifference}, \texttt{faceAngleDifference}, and \texttt{dtwValues}: Centers features by removing the mean and scaling to unit variance, which is suitable for normally distributed features.
\end{itemize}

\begin{table}[H]
    \centering
    \resizebox{\textwidth}{!}{
    \begin{tabular}{@{}llp{0.6\textwidth}@{}}
    \toprule
    \textbf{Input (Features)} & \textbf{Feature Name} & \textbf{Description} \\ 
    \midrule
    1 & \texttt{interDistance} & Euclidean distance between the two pedestrians in the pair. \\
    2 & \texttt{timeDifference} & Absolute difference in timestamps between the two pedestrians in the pair. \\
    3 & \texttt{velocityDifference} & Absolute difference in velocity between the two pedestrians in the pair. \\
    4 & \texttt{motionAngleDifference} & Absolute difference in motion angles between the two pedestrians in the pair. \\    
    5 & \texttt{faceAngleDifference} & Absolute difference in facing angles between the two pedestrians in the pair. \\
    6 & \texttt{dtwValues} & Pair trajectory similarity metric calculated using Dynamic Time Warping (DTW). \\
    \bottomrule
    \textbf{Output (Label)} & \texttt{label} & Predicted label indicating whether the pair is a flock (1) or not (0). \\ 
    \bottomrule
    \end{tabular}
    }    
    \caption{Feature Description for Binary Classification Sequential Models (Inputs and Output)}
    \label{tab:feature_table}
\end{table}

\subsubsection*{Binary Model Parameter Selection}

In the process of selecting the optimal hyperparameters for our binary classification sequential model, we primarily draw from established best practices in the field. The model's hyperparameters, including batch size and hidden layer sizes, play a critical role in ensuring effective learning and generalization. We initially follow guidelines found in key research papers, which focus on the importance of carefully tuning these parameters to achieve robust performance.

For batch sizes and hidden layer sizes, we refer to studies such as \cite{deepsurrogate} and \cite{batchsizepaper}, which explore the impact of various batch sizes on model stability and convergence rates. Another important paper is \cite{hiddenlayers}, which discusses the effects of hidden layer sizes and their impact on model expressiveness and training efficiency.

To select the best-performing hyperparameters, we conduct tests across different combinations of batch sizes and hidden layer sizes. Specifically, we test the following combinations:
\begin{itemize}
    \item \textbf{Batch Sizes}: [64, 32, 16, 8]
    \item \textbf{Hidden Layer Sizes}: [256, 128, 64, 32, 16]
\end{itemize}

These combinations are chosen to balance the trade-off between computational efficiency and model performance. Larger batch sizes may speed up training but could lead to poorer generalization, while smaller batch sizes might improve generalization at the cost of longer training times. Similarly, different hidden layer sizes can affect the model's ability to capture complex patterns in the data while maintaining computational feasibility.

The result sections will describe the results from conducting experiments with these parameter combinations, comparing their performance based on validation accuracy and other relevant metrics.

\subsubsection{Multi-agent group data preparation}

As described in Section \ref{Group_Data_Preperation}, we prepare timestamp bin data by aligning the starting point of each pedestrian within the current timestamp range.

\begin{figure}[H] 
    \centering
    \ifpdf
        \includegraphics[width=0.75\textwidth]{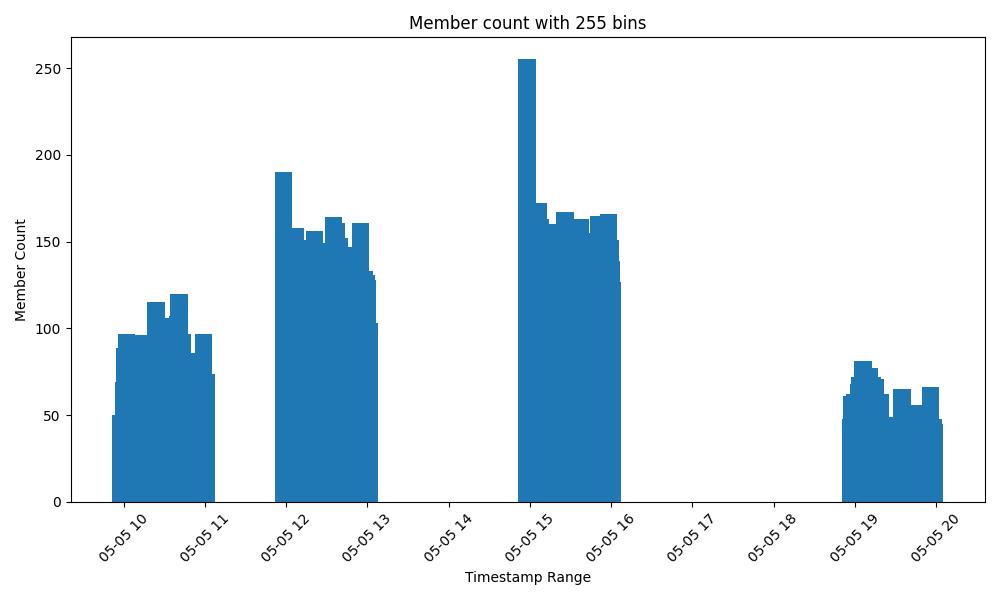}  % Use PDFLaTeX-compatible formats
    \fi
    \caption{Sample Member Count for Each Bin on 2013/05/05}
    \label{fig:member_count_chart}
\end{figure}

The bar chart shown in Figure \ref{fig:member_count_chart} visualizes the \textbf{member count} across distinct \textbf{time ranges} from the dataset. The data is grouped into four primary time periods that represent significant hours of pedestrian activity, as described in \url{https://dil.atr.jp/sets/groups/}:

\begin{itemize} \item 10:00 - 11:00 \item 12:00 - 13:00 \item 15:00 - 16:00 \item 19:00 - 20:00 \end{itemize}

Each time range is subdivided into 1-minute intervals, with the height of the bars indicating the number of pedestrians observed in each interval. The distribution of pedestrians across these time ranges allows for a more detailed analysis of pedestrian activity within each hour. Given that the sequence length used in the analysis was 100, the data is processed in 1-minute bins to capture a finer temporal resolution of pedestrian movement.

These bins provide the prepared scene units for subsequent offline pairwise evaluation and flock identification, supporting the analysis of collective pedestrian dynamics. 

Depending on the duration of the time interval and the sequence length of the sample, the number of bins and the member count in each bin may vary. In this study, we set the sequence length to 100 and the time interval to 1-minute for the experiment.

\begin{figure}[H]
    \centering
    % First row of images
    \ifpdf
        \begin{subfigure}[c]{0.45\textwidth} 
            \centering
            \includegraphics[width=\textwidth]{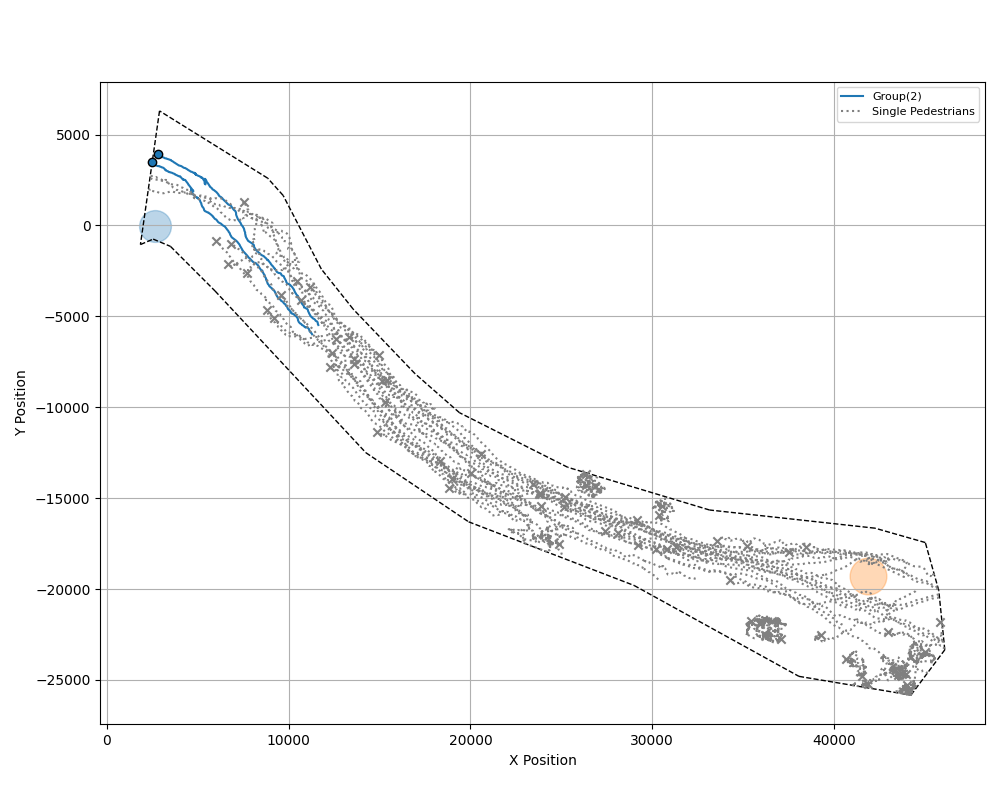}
            \caption{Sample Bin Containing Few Groups.}
            \label{fig:Scene1}
        \end{subfigure}
    \fi
    \hfill
    \begin{subfigure}[c]{0.45\textwidth}
        \centering
        \includegraphics[width=\textwidth]{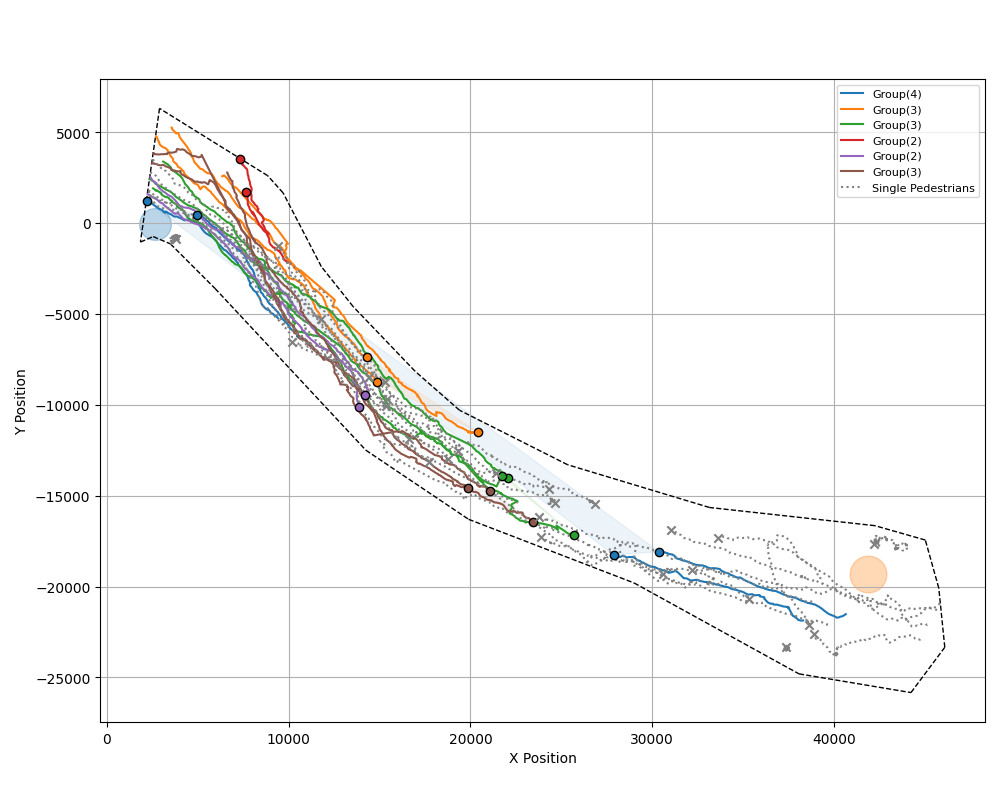}
        \caption{Sample Bin Containing Predominantly Groups.}
        \label{fig:scene2}
    \end{subfigure}

    \caption{Overview of Sample Bins Containing Group and Single Pedestrians.}
    \label{fig:sample_bin}
\end{figure}

In Figure~\ref{fig:sample_bin}, the sample bins illustrate the distribution of group and single pedestrians used as input for the flock detection process. Each bin contains trajectories segmented over a given time window, and the algorithm operates by performing pairwise comparisons within these bins to evaluate spatial and temporal proximity. Based on these comparisons, the model predicts whether each pedestrian pair forms part of a flock, subsequently aggregating positive pairwise predictions to identify complete flocks. This representation highlights both mixed cases—where groups and singles coexist—and bins dominated by either few or many group pedestrians.

\subsection{Computational Environment}

The computational environment for the present study consists of a high-performance workstation equipped with an Intel(R) Core(TM) i9-13900 processor (32 CPUs), 64 GB of RAM, and a powerful 32GB NVIDIA GeForce RTX 4070 Ti GPU. The system is configured to support intensive computational tasks and is running on Python 3.12. For deep learning tasks, we leverage PyTorch, with TabNet-4.1 being employed for regression and classification tasks.

The following software libraries and frameworks were utilized in this study:

\begin{itemize}
    \item \textbf{Deep Learning Models}: We implemented various models using \textit{PyTorch}, including Recurrent Neural Networks (RNN), Long Short-Term Memory networks (LSTM), and Transformer-based models. Key hyperparameters include \texttt{learning\_rate} = 0.001 and \texttt{num\_epochs} = 1000.
\end{itemize}

\begin{itemize}
    \item \textbf{Similarity Measurement}: For calculating the similarity between trajectories, we utilized the \textit{fastdtw} package, which efficiently computes the Euclidean distance between time-series data.
\end{itemize}

\begin{itemize}
    \item \textbf{Visualization}: We used the \textit{Matplotlib} library for high-quality visualizations of the flocking patterns, allowing for clear and efficient presentation of the agent trajectories and group dynamics.
\end{itemize}

\subsection{Main Results}

In this section, we present the results of the binary classification model by evaluating and comparing the performance of RNN, LSTM, and Transformer models in flock detection. The experiments were conducted on multiple datasets with varying sequence lengths (i.e., number of samples) to assess the impact of sequence size on model performance. The results indicate that both runtime and classification accuracy vary significantly depending on the model architecture, even when applied to the same dataset.

\subsubsection{Binary Classification Accuracy by Data Length and Models}

The following reported accuracy values represent the \textbf{average of 20-independent runs} (combination of batch  and hidden layer size) for each model. The results highlight the increasing performance of all three models (RNN, LSTM, Transformer) as sequence length increases.

\begin{figure}[H] 
    \centering
    \ifpdf
        \includegraphics[width=0.75\textwidth]{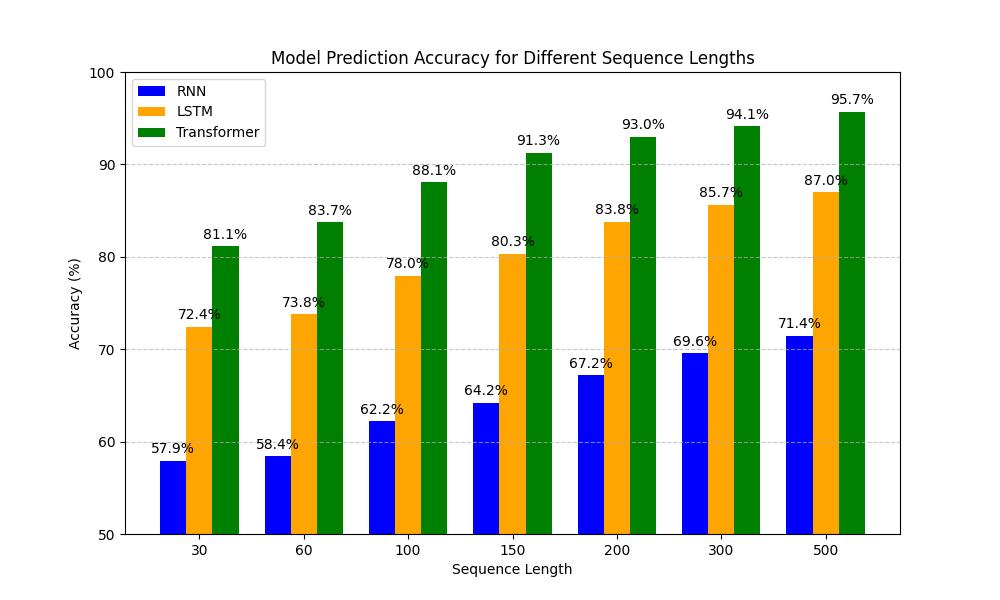}  % Use PDFLaTeX-compatible formats
    \fi
    \caption{Sequential models prediction accuracy for different sequences}
    \label{fig:model_accuracy}
\end{figure}

\begin{itemize}
    \item Transformer Model Performance. 
    The Transformer consistently outperforms both RNN and LSTM across all sequence lengths. Notably, at \textbf{sequence length = 100}, the Transformer model exceeds \textbf{90\% accuracy} in some runs, averaging \textbf{88.06\%}. It continues to improve with longer sequences, reaching \textbf{95.70\% at sequence length = 500}.
    \item LSTM Model Performance.
    The LSTM model demonstrates \textbf{strong performance for longer sequences}, surpassing \textbf{80\% accuracy} at \textbf{150 sequences} and achieving \textbf{86.96\% at 500 sequences}. It consistently outperforms RNN across all sequence lengths, showing its ability to better capture long-term dependencies.
    \item RNN Model Performance.
    The RNN model lags behind the LSTM and Transformer, especially for longer sequences. While it shows improvement with increased sequence length, it struggles to exceed \textbf{70\% accuracy} even at \textbf{500 sequences}, peaking at \textbf{71.43\%}.

\end{itemize}

\textbf{Key Observations}
\begin{itemize}
    \item \textbf{Longer sequences benefit all models}, with the Transformer making the most significant gains.
    \item \textbf{LSTM handles longer sequences better than RNN}, suggesting that its gating mechanisms effectively capture dependencies.
    \item \textbf{The Transformer is the best-performing model} across all tested sequence lengths, achieving over \textbf{90\% accuracy} for sequences \textbf{$\geq$ 150}.
\end{itemize}

\subsubsection{Model Training Time Comparison}

The reported runtime values represent the \textbf{average of 20 independent runs}, considering different combinations of batch size and hidden layer size for each model. The results highlight the increasing runtime of all three models (RNN, LSTM, Transformer) as sequence length increases. Since actual training involved early stopping, the runtime could vary across runs. To ensure a fair comparison of model training efficiency, we set the number of epochs to 1000 for all training instances. 

\begin{figure}[H] 
    \centering
    \ifpdf
        \includegraphics[width=0.75\textwidth]{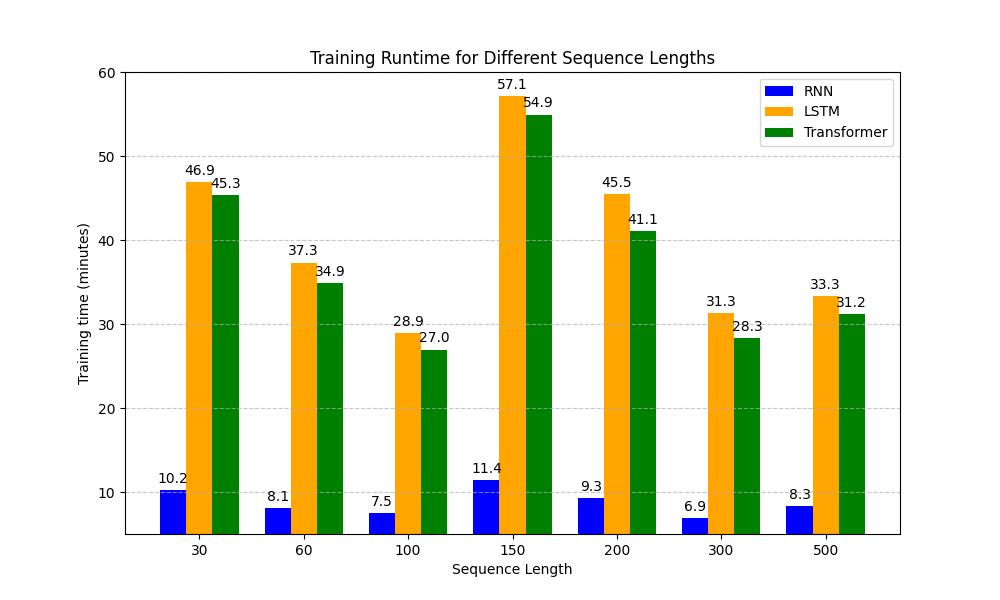}  % Use PDFLaTeX-compatible formats
    \fi
    \caption{Model training time for different sequences}
    \label{fig:model_training_runtime}
\end{figure}

The runtime comparison results, shown in Fig.~\ref{fig:model_training_runtime}, indicate that the \textbf{RNN} model consistently achieves the shortest training times across all tested sequence lengths, making it the most computationally efficient among the three architectures. This performance advantage is largely attributable to its relatively simple recurrent structure, which requires fewer parameters and operations per time step.

In contrast, the \textbf{LSTM} model exhibits the longest training times for all sequence lengths, with a peak runtime of 57.1 minutes at a sequence length of 150. This overhead is likely caused by the additional gating and memory cell computations inherent in LSTM’s design, which, while beneficial for long-term dependency modeling, substantially increase computational complexity. The \textbf{Transformer} model’s runtimes are generally shorter than LSTM’s but remain higher than RNN’s, reflecting its reliance on multi-head attention and feed-forward blocks, which scale with sequence length.

An additional observation is that training time does not monotonically increase with sequence length. Instead, the fastest runtimes occur at sequence lengths of 100 and 300 across all models, while the longest runtimes are observed at sequence length 150. This suggests the presence of computational inefficiencies at certain sequence lengths, potentially arising from hardware-level memory allocation or parallelization behavior during training.

\textbf{Key Observations}
\begin{itemize}
\item \textbf{RNN} is the fastest model across all sequence lengths due to its simpler architecture and lower parameter count.
\item \textbf{LSTM} incurs the highest training cost, peaking at 57.1 minutes for sequence length 150, owing to its complex gating mechanisms.
\item \textbf{Transformer} runtimes are generally shorter than LSTM but longer than RNN, reflecting its attention-based design.
\item Sequence lengths of 100 and 300 yield the shortest runtimes across all models, while 150 results in the longest runtimes.
\end{itemize}

These findings highlight a trade-off between model complexity and training efficiency. While LSTMs and Transformers are generally more expressive for sequence learning tasks, they require significantly longer training times compared to RNNs. Furthermore, sequence length selection can influence training efficiency in non-linear ways, suggesting that computational profiling should be considered alongside predictive performance when choosing model configurations.

\subsubsection{Comparison of Accuracy and Runtime Across Models}

\begin{figure}[H] 
    \centering
    \ifpdf
        \includegraphics[width=0.75\textwidth]{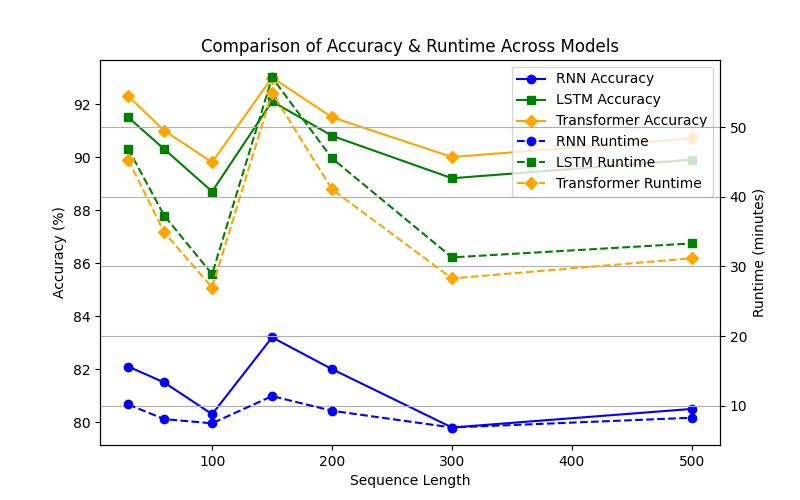}  % Use PDFLaTeX-compatible formats
    \fi
    \caption{Model training time for different sequences}
    \label{fig:model_training_accuracy_runtime}
\end{figure}

The figure illustrates the trade-off between \textbf{accuracy (\%)} and \textbf{runtime (minutes)} for RNN, LSTM, and Transformer models across different sequence lengths.

\begin{itemize}
    \item \textbf{Accuracy:} Transformer achieves the highest accuracy, followed by LSTM, while RNN performs the worst.
    \item \textbf{Runtime:} RNN is the fastest, while LSTM takes the longest time. Transformer is slightly faster than LSTM.
    \item \textbf{Sequence Length:} Training is more efficient at sequence lengths \textbf{100 and 300}, while \textbf{150 takes the longest}.
    \item \textbf{Key Trade-off:} RNN is best for speed, while Transformer offers a better balance between accuracy and runtime.
\end{itemize}

\subsubsection{Sample Screens for pair prediction}

Below are sample pair detection results for a sequence length of 100. The model accurately predicts complex trajectories as pairs in many cases. However, in some instances, it may incorrectly detect two individual pedestrians moving in similar directions as a pair.

\begin{figure}[H]
    \centering
    % First row of images
    \ifpdf
        \begin{subfigure}[c]{0.45\textwidth} 
            \centering
            \includegraphics[width=\textwidth]{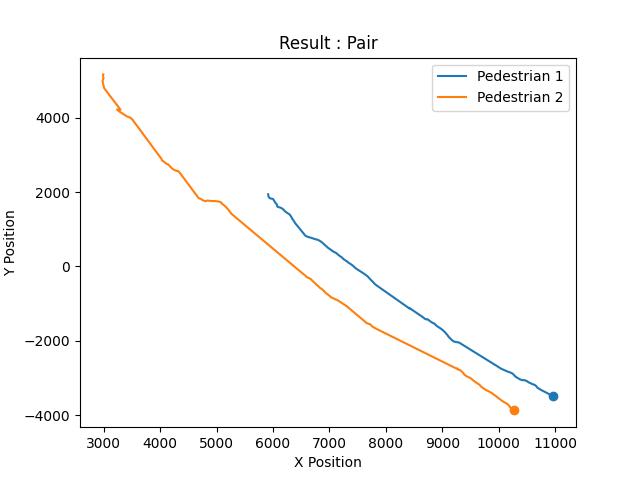}
            \caption{The pair prediction result for straight trajectories.}
            \label{fig:pair1b}
        \end{subfigure}
    \fi
    \hfill
    \begin{subfigure}[c]{0.45\textwidth}
        \centering
        \includegraphics[width=\textwidth]{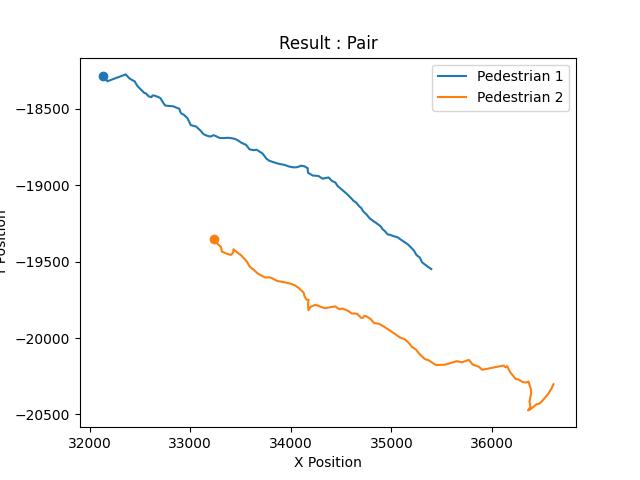}
        \caption{The pair prediction result for complex trajectories.}
        \label{fig:pair2}
    \end{subfigure}

    \caption{Sample pair detection results for a sequence length of 100.}
    \label{fig:pair_prediction}
\end{figure}

\subsubsection{Flock Validation}

We validate our flock detection algorithm using the group files described in \cite{zanlungo2015pedestrian}. These group files contain group-labeled data for 6 days. After performing data preprocessing and cleaning, we found that most flocks have sizes of 2, 3, at the most between 11 and 12 across all the group files. We validated our method by training the model on data from different days, introducing unseen data to assess the model's generalization ability.

As we described in \ref{multi-agent} our flock detection algorithm iterates through all combinations to predict pairs and then filters the results to determine the flock size based on pair predictions. 

We further describe an example of flock validation results for the groups on 2013/02/17, using a trained Transformer model with a sequence length = 100, batch size = 32, hidden layer size=64 and model accuracy 91.35\%. We set the confidence score for pair prediction to 0.9. As a result, all the flocks were correctly detected, with the following "group size":"count" distribution:

\textit{{"2": 1528, "4": 143, "3": 448, "5": 36, "7": 4, "6": 6, "11": 1, "9": 1, "8": 2}}

\subsubsection{Sample Screens for flock detection}

\begin{figure}[H]
    \centering
    % First row of images
    \ifpdf
        \begin{subfigure}[c]{0.45\textwidth} 
            \centering
            \includegraphics[width=\textwidth]{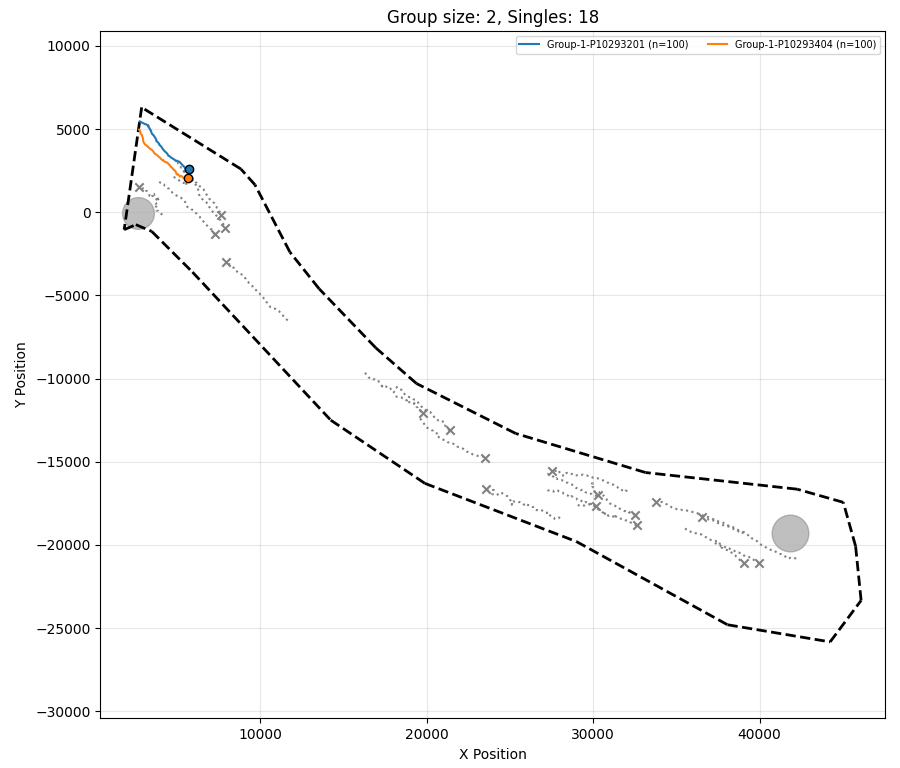}
            \caption{Detected flock size with 2.}
            \label{fig:flock2}
        \end{subfigure}
    \fi
    \hfill
    \begin{subfigure}[c]{0.45\textwidth}
        \centering
        \includegraphics[width=\textwidth]{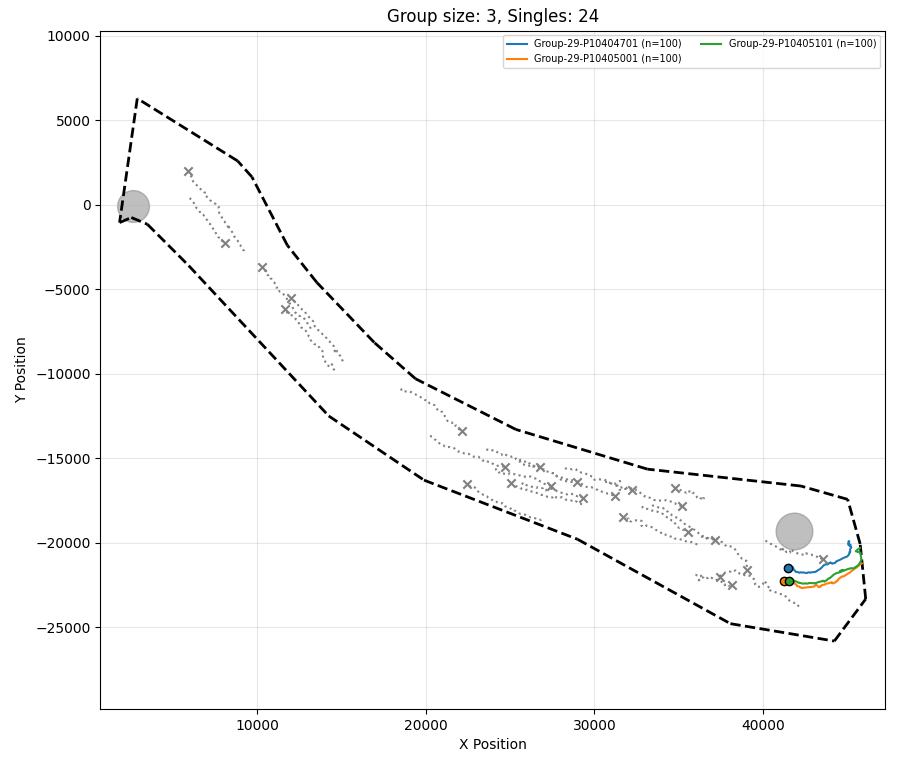}
        \caption{Detected flock size with 3.}
        \label{fig:flock3}
    \end{subfigure}

    % Second row of images
    \begin{subfigure}[c]{0.45\textwidth}
        \centering
        \includegraphics[width=\textwidth]{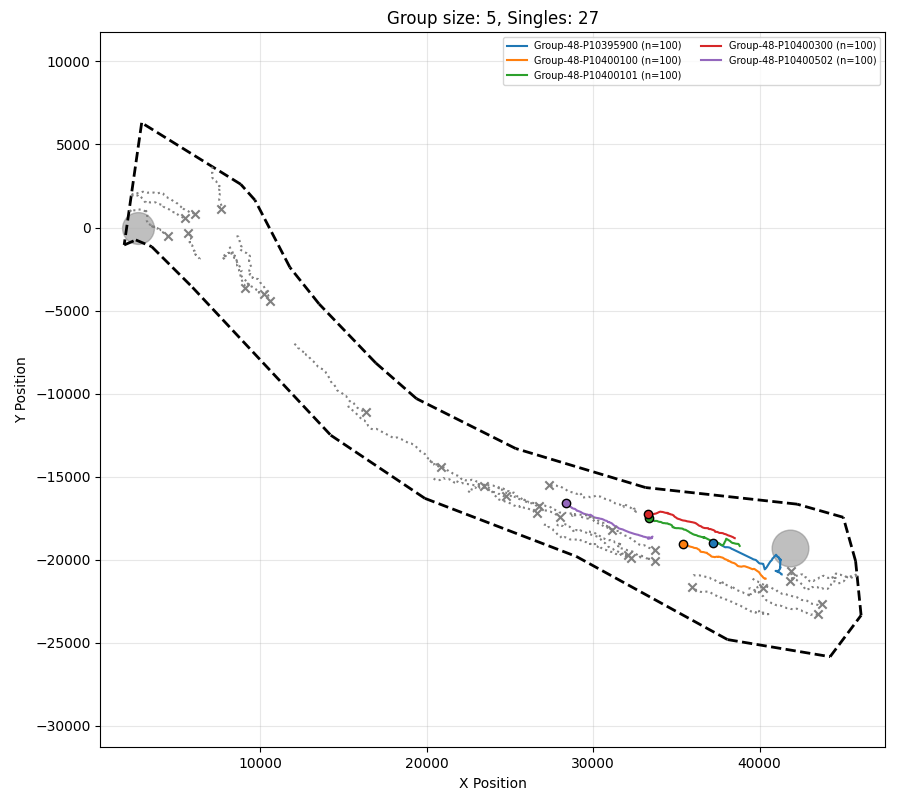}
        \caption{Detected flock size with 5.}
        \label{fig:flock5}
    \end{subfigure}
    \hfill
    \begin{subfigure}[c]{0.45\textwidth}
        \centering
        \includegraphics[width=\textwidth]{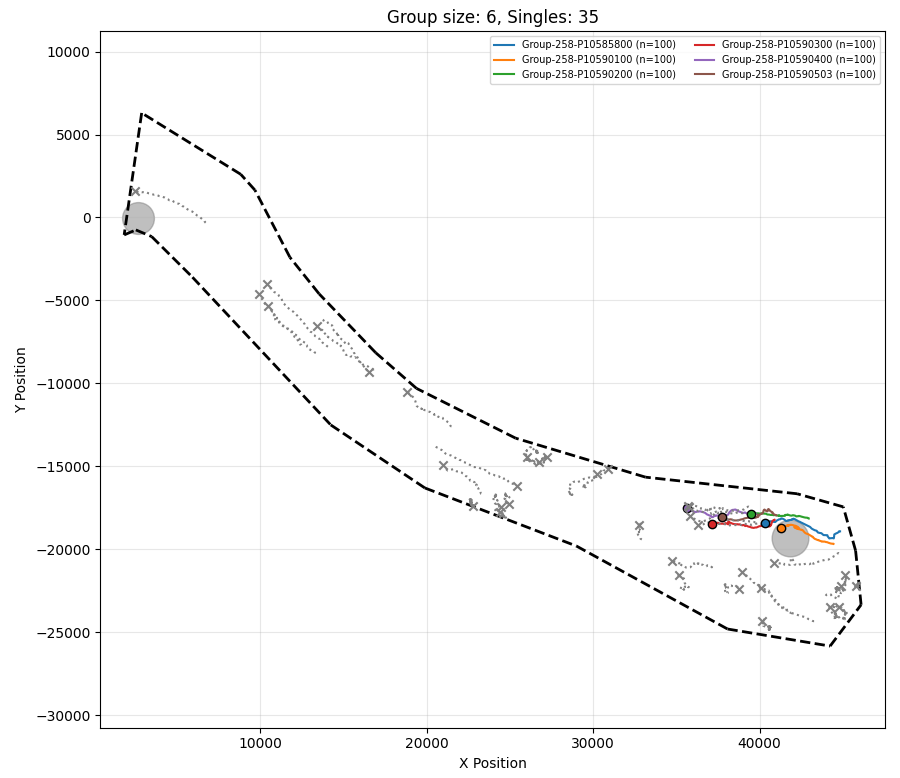}
        \caption{Detected flock size with 6.}
        \label{fig:flock6}
    \end{subfigure}

    \caption{Sample Flock Detection Results for Four Different Flock Sizes.}
    \label{fig:flock_detection}
\end{figure}

\section{Discussion}

In this section, we discuss the results of our flock detection method and highlight key observations regarding the impact of data sequence length and the dependency of features used in the model.

\subsection*{Impact of Data Sequence Length}
The performance of our model demonstrates that using a sequence length of 100 steps results in over 90\% accuracy for pair detection with a set of 8 features. While this sequence length yields high accuracy, it is worth noting that 100 steps might seem relatively long from a human perspective when identifying pairs in a trajectory. Human perception of trajectories typically involves recognizing patterns and interactions over shorter durations. Despite this, our results indicate that using longer sequences provides the model with more comprehensive information about the agents' movements, which is crucial for accurately predicting pairings. However, a balance between sequence length and model performance must be considered, as excessively long sequences could lead to increased computational complexity and diminishing returns in predictive accuracy beyond a certain point.

\subsection*{Feature Dependency}
The current feature set for the binary classification model, consisting of inter-agent distance, timestamp difference, velocity difference, motion-angle difference, facing-angle difference, and DTW-based trajectory similarity, has shown promising results in detecting pedestrian pairs. In some experiments, different feature combinations yielded similar accuracy. This suggests that the model can identify pairs using a subset of the spatio-temporal features, although further controlled feature-ablation experiments would be needed to establish the contribution of each feature. Reducing the feature set may improve computational efficiency while maintaining classification performance. The runtime measurements in this study concern the reported experimental processing and model configurations; they do not establish end-to-end real-time capability.

In both of these cases, we further analyze and conduct additional experiments to explore the optimal balance between sequence length and feature set for improving model efficiency and generalization across various scenarios.

\subsection*{Scope and Limitations}
The workflow evaluated in this study is an offline analysis pipeline for recorded pedestrian trajectories. Data preparation, time-bin grouping, scene generation, and model training are carried out before the trained classifier is applied to prepared scenes. Although the model performs pairwise classification over trajectory sequences and the resulting predictions are used to identify flocks, the experiments do not evaluate continuous processing of live data. In particular, no end-to-end latency, online update strategy, or real-time deployment performance is reported. Consequently, the results should be interpreted as evidence for offline moving-flock identification rather than as a demonstration of a real-time flock-detection system.

\section{Conclusion}

We proposed a moving-flock identification workflow that uses a trained binary classifier to classify pedestrian trajectory pairs and aggregates positive pairwise predictions into multi-agent groups. The workflow accommodates prepared scenes with varying numbers of pedestrians and can therefore be used to analyze recorded pedestrian movement under different conditions. The experiments evaluate the classification models and the subsequent flock-identification procedure on prepared trajectory data. They do not demonstrate an end-to-end real-time system, since trajectory preprocessing, scene generation, and model training are performed in advance and the evaluation is conducted on prepared scenes.

Additionally, the method demonstrates versatility by extending to more complex group behaviors, such as the identification of convoys or swarms. This enhancement is particularly valuable in scenarios where agent interactions form intricate patterns that need to be understood for effective crowd control or behavioral analysis.

Our current experiments use a sequence length of 100 for the reported flock-validation example. Further work will investigate how sequence length affects classification accuracy, computational cost, and the temporal coverage required to characterize a flock. Shorter sequences may reduce processing requirements, while longer sequences may provide more movement context. Any future application to live or streaming data would require a separate system design and evaluation, including online data acquisition, incremental preprocessing, inference latency, and end-to-end throughput; these aspects are outside the scope of the present study.

Furthermore, the method can be enhanced by experimenting with different sequential deep learning models, such as bidirectional LSTMs or attention-based models like the Transformer. These models could potentially improve accuracy in identifying group dynamics by leveraging both past and future trajectory data. Exploring different model architectures and incorporating hybrid models could also enhance the system’s adaptability to various environments and increase its ability to generalize across unseen data. In addition to sequence length optimization, the integration of more sophisticated model structures may further contribute to the development of scalable and efficient flock detection methods.

In summary, the experiments show the potential of sequential deep learning models for pairwise flock classification on prepared pedestrian trajectory data. Future work will examine model architectures, feature selection, sequence-length choices, and generalization across additional datasets. A real-time or online deployment would require additional development and evaluation beyond the offline workflow studied here.

\bibliographystyle{unsrt}
\bibliography{references}

\end{document}